\def\year{2020}\relax
\documentclass[letterpaper]{article} 
\usepackage{aaai20}  
\usepackage{times}  
\usepackage{helvet} 
\usepackage{courier}  
\usepackage[hyphens]{url}  
\usepackage{amsmath,amssymb,amsfonts}
\usepackage{adjustbox,lipsum}
\usepackage{blindtext}
\usepackage{graphicx} 
\usepackage{graphicx}  
\usepackage{xcolor}
\usepackage{verbatim}

\usepackage{todonotes}

\title{Detecting Asks in Social Engineering Attacks:\\ Impact of Linguistic and Structural Knowledge}

\author{Bonnie J. Dorr,\textsuperscript{\rm 1}
Archna Bhatia,\textsuperscript{\rm 1} 
Adam Dalton,\textsuperscript{\rm 1}
Brodie Mather,\textsuperscript{\rm 1}\\ 
\Large \textbf{
Bryanna Hebenstreit,\textsuperscript{\rm 2}
Sashank Santhanam,\textsuperscript{\rm 3}
Zhuo Cheng,\textsuperscript{\rm 3}}\\
\Large \textbf{
Samira Shaikh,\textsuperscript{\rm 3}
Alan Zemel,\textsuperscript{\rm 2}
Tomek Strzalkowski\textsuperscript{\rm 4}}\\
\textsuperscript{\rm 1}Institute for Human and Machine Cognition (IHMC), Ocala, FL, USA, \{bdorr,abhatia,adalton,bmather\}@ihmc.us\\ 
\textsuperscript{\rm 2}State University of New York, Albany, NY, USA, \{bhebenstreit,azemel\}@albany.edu\\ 
\textsuperscript{\rm 3}University of North Carolina, Charlotte, NC, USA, \{sshaikh2,ssantha1,zcheng5\}@uncc.edu\\ 
\textsuperscript{\rm 4}Rensselaer Polytechnic Institute, Troy, NY, USA, tomek@rpi.edu
}

\begin{document}

\maketitle

\begin{abstract}
Social engineers attempt to manipulate users into undertaking actions such as downloading malware by clicking links or providing access to money or sensitive information.  Natural language processing, computational sociolinguistics, and media-specific structural clues provide a means for detecting both the \textit{ask} (e.g., \textit{buy gift card}) and the risk/reward implied by the ask, which we call \textit{framing} (e.g., \textit{lose your job}, \textit{get a raise}). We apply linguistic resources such as \textit{Lexical Conceptual Structure} to tackle ask detection and also leverage structural clues such as links and their proximity to identified asks to improve confidence in our results. Our experiments indicate that the performance of ask detection, framing detection, and identification of the \textit{top ask} is improved by linguistically motivated classes coupled with structural clues such as links.  Our approach is implemented in a system that informs users about social engineering risk situations.
\end{abstract}

\section{Introduction}
Social engineering (SE) attacks are a significant cybersecurity threat, putting individuals and organizations at risk. 
Social engineers attempt to manipulate users into undertaking actions such as downloading malware by clicking links (PERFORM) or providing access to money or sensitive information (GIVE). Such eliciting behaviors, or \textit{asks}, are not always explicitly stated in text \cite{Drew2014Requesting}, but may be gleaned through automatic techniques that employ both linguistic and structural knowledge. Natural language processing (NLP), computational sociolinguistics, and media-specific structural clues provide a means for detecting both the \textit{ask} (e.g., \textit{buy gift card}) and the risk/reward (or LOSE/GAIN) implied by the ask, which we call \textit{framing} (e.g., \textit{lose your job}, \textit{get a raise}). These elements can be used in downstream operations for countering attacks, e.g., through bot-produced responses and actions.

Interest in conversational intelligence has surged in recent years as evidenced by the ConvAI2 NeurIPS competition \cite{Dinan:2019,yusupov-kuratov-2018-nips} and related efforts in natural language dialogue
\cite{Perera:2017}. We build on prior Cyber-related dialogue work \cite{Dalton:2019}, focusing on critical conversational objectives beyond NLP-based cyber-attack \textit{detection} and \textit{prediction} \cite{dalton2017BDA4CID,Perera:2018,Hollingshead:2019} or NLP-based \textit{generation} of warnings and the explanations behind them \cite{Kazakova:2019}. We implement a novel ask detection approach with the ultimate goal of sustaining a believable \textit{conversation} with a social engineer to extract their true identity without putting valuable resources at risk.

Toward that end, we apply linguistic resources, including \textit{Lexical Conceptual Structure} (e.g., relating verbs such as \textit{give}, \textit{donate}) \cite{Dor:18a,dorr-voss:2018} and \textit{categorial variation} (e.g., relating word variants such as \textit{reference}, \textit{refer}) \cite{Nizar:2003}.  We also leverage structural clues such as links (URLs, email addresses) and their proximity to identified asks. Although the email channel is our starting point, we adopt such techniques with an eye toward extensibility to channels beyond email (e.g., texting), but at the same time we identify techniques specific to email, such as email signature and quoted reply removal. 
No training data are needed for the approach described herein.

\begin{table}
    \centering
    \begin{tabular}{|p{1.2in}|p{.48in}|p{.62in}|r|}\hline
     \textbf{ Email}  & \textbf{Framing} & \textbf{Ask} & \textbf{Conf} \\ \hline
      I am stuck at the airport. Please help me out by sending \$500. 
      & LOSE\newline stuck() \newline [] 
      & PERFORM help() \newline [finance \_money] 
      &  0.8 \\ \hline
      
      It is a pleasure to inform you that you have won \$1.5M. Contact me. \newline ({\color{blue}\underline{jw11@example.com}})
      & GAIN \newline won()\newline [finance \_money] 
      & PERFORM contact \newline ({\color{blue}\underline{jw11@...}}) \newline [] 
      & 0.9 \\ \hline

      Your dog could win prizes. {\color{blue}\underline{Vote now}}.
      &GAIN \newline win()\newline []
      &PERFORM \newline vote \newline({\color{blue}\underline{http...}}) []
      &0.9\\ \hline

      After you submit, we will pick finalists in each category. Users will vote on their favorite three winners. 
      & GAIN \newline pick()\newline [] 
      & GIVE \newline vote() \newline [] 
      & 0.6 \\ \hline

    \end{tabular}
    \caption{Representative system output for four emails: framings, asks, and confidence scores. Parentheses () designate clickable links (emails, URLs,...) and brackets [] indicate the ask/framing type of interest to the social engineer.}
    \label{tab:Examples}
\end{table}

Table~\ref{tab:Examples} shows representative system output 
produced by our approach for four (presumed) SE emails.  \textit{Framing} output sets the stage for the ask, i.e., the purported threat (LOSE) or benefit (GAIN) that the social engineer wants the potential victim to believe is dependent on compliance or lack thereof. The output shows three PERFORM asks and one GIVE ask. Information in parentheses () refers to one or more links that the potential victim might choose to click.\footnote{For space reasons, abbreviated forms ``jw11@...'' and ``http...'' are used in the table.} Links are detectable from html mark-up whether or not their textual form is used in the email body. Information in brackets [] refers to the ask/framing category of interest to the social engineer (e.g., finance\_money, personal, credentials).\footnote{Experiments reported herein do not focus on such categories, but these are shown for illustrational purposes. (Future work will investigate the impact of such categories on response generation.)}

We describe multiple experiments to determine the impact of linguistic and structural knowledge on ask/framing detection. Two key findings are: (1) Both linguistically motivated classes (e.g., \textit{Judge, Want, Send)} and verbal processing (e.g., filtering on the basis of part of speech) improve precision and recall of ask and framing detection; 
(2) Structural clues such as links provide a fuller context for asks (e.g., \textit{click here}) and improve precision and recall of ``top ask'' detection (i.e., identification of critical asks for downstream response generation) from individual emails. These asks are then converted to a threat-intelligence representation where they make up the matchable pattern. Our approach is implemented and currently under evaluation by prominent transition partners. An accompanying risk model ties the target, and what is being asked of them, to an attack signature that can be mapped to a threat actor; the mapping then reveals other stages of the attack and what the motivation might be. 

The next section presents the foundational background for asks and framing. We subsequently present our approach, focusing on the linguistic and structural knowledge adopted into our solution and describing our algorithm with representative examples. We present a range of experiments and results and discuss the upshot of our experiments and present related work, contrasting prior approaches to our own. We conclude with ideas for future work. By-products of this study are available at a website henceforth referred to as \textit{Ask Detection} webpage, for a larger project called PANACEA: \url{https://social-threats.github.io/panacea-ask-detection/}.

\section{Foundational Elements: Asks and Framing}

The notions of \textit{ask} and \textit{framing} are defined as they pertain to our task of ask/framing detection in the SE context.

\subsection{What is an Ask?}

The notion of an \textit{ask} is closely related to the notion of a request \cite{Zemel2017Texts}. Four related phenomena are subsumed under the heading of an \textit{ask}: a) \textit{proposal}, b) \textit{offer}, c) \textit{request} and d) \textit{suggestion} \cite{Couper2014Grammar}. An \textit{ask} is an action that ``should be taken broadly to include other ways of asking than speaking'' \cite{Drew2014Recruitment}. 
The nature of the action requested will influence both the form of the \textit{ask} and the nature of the expected and actual response to the \textit{ask} \cite{Drew2014Recruitment}. 
We identify two ask types: a
topical descriptor that specifies the kind of thing being asked for (GIVE), and an action associated with its delivery or production (PERFORM). 
In SE attacks, asks are routinely used to recruit recipients to perform actions that provide money, information or system access.  

Perhaps most importantly, an ask 
elicits relevant  responses from the recipient,
thus providing an
opportunity to 
generate responses 
that
elicit information about the attacker from the attacker. 
The same social obligation to respond to a request that a social engineer uses to elicit responses
can be used to elicit information about the social engineer.  

We have conducted pilot experiments with both staged and ``live'' social engineering attacks and noted that general conversational norms are observed, as expected. While one or both sides of the conversation engage in deception, they still need to accomplish their objectives, e.g., keeping the conversation alive, and obtaining information they want. Correspondingly, our approach treats ask/framing combinations
as sense-making practices. All actors rely on shared knowledge of, and shared ability to recognize, how asks and framings fit together in conventional ways to produce meaningful interactions \cite{GarfinkelSacks1970,PomerantzFehr2011,pomerantz2017inferring,Sacks1992,Schegloff2007}. We rely on this understanding of communicative practice in designing our system to both recognize and implement legitimate and deceptive messaging.

\subsection{What is Framing?}

\textit{Framing} 
refers to 
linguistic and social resources used to persuade the recipient of an ask to comply and perform the requested social action. An ask creates a social obligation to respond, but does not necessarily provide an adequate 
basis for compliance with the ask. An \textit{ask} must be ``contextualized'' to be persuasive \cite{Huma2018Persuasive}. Recognizing and producing \textit{framing} thus significantly shapes the kind of response the \textit{ask} makes relevant. 

Framing and the particulars of ask-construction 
enable creation of a
``benefactive stance'' \cite{Clayman2014Benefactor} that motivates compliance and also
enables and constrains
the ways that recipients respond \cite{Huma2018Persuasive}. 
This may involve 
asking a question that makes production of an answer conditionally relevant. Alternatively, the benefit 
or cost 
that purportedly may accrue to the recipient (based on 
compliance or non-compliance) provides the basis upon which performance of the \textit{ask} can be decided. Framing consists of just these specifications of benefit (GAIN) and cost (LOSE). 

Social engineers
rely
on the fact that we routinely (often without much consideration) conform to 
social interaction conventions based on the credibility of the framing used. The efficacy of framing is not limited only to victims of SE attacks. Asks and framing directed at social engineers make them 
targets of counter-attacks. 

\section{Approach to Ask/Framing Detection}

This section describes our approach to ask/framing detection, starting with the application of linguistic knowledge and structural knowledge, and finishing with a description of the algorithmic steps and confidence score. 

\subsection{Application of Linguistic Knowledge}

\begin{figure}
\begin{tabular}{|l|}\hline
\textbf{a.~Dependency:}\\
\includegraphics*[width=3.1in]{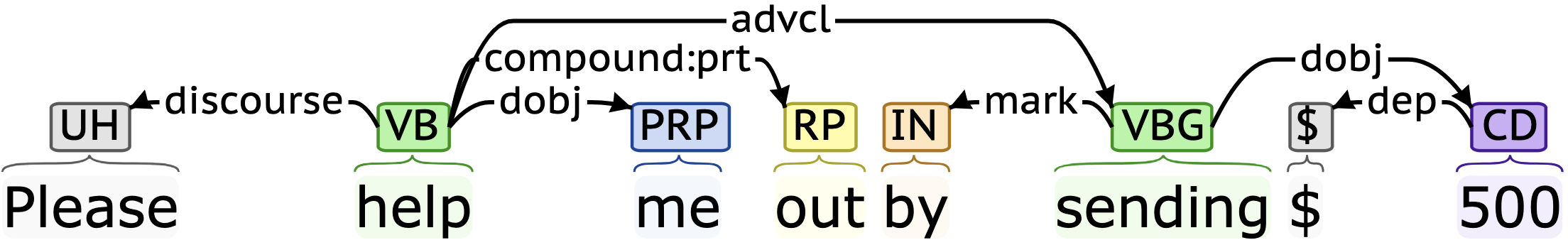}\\ \hline
\textbf{b.~Constituency:}\\
\includegraphics*[width=3.1in,height=2.5in]{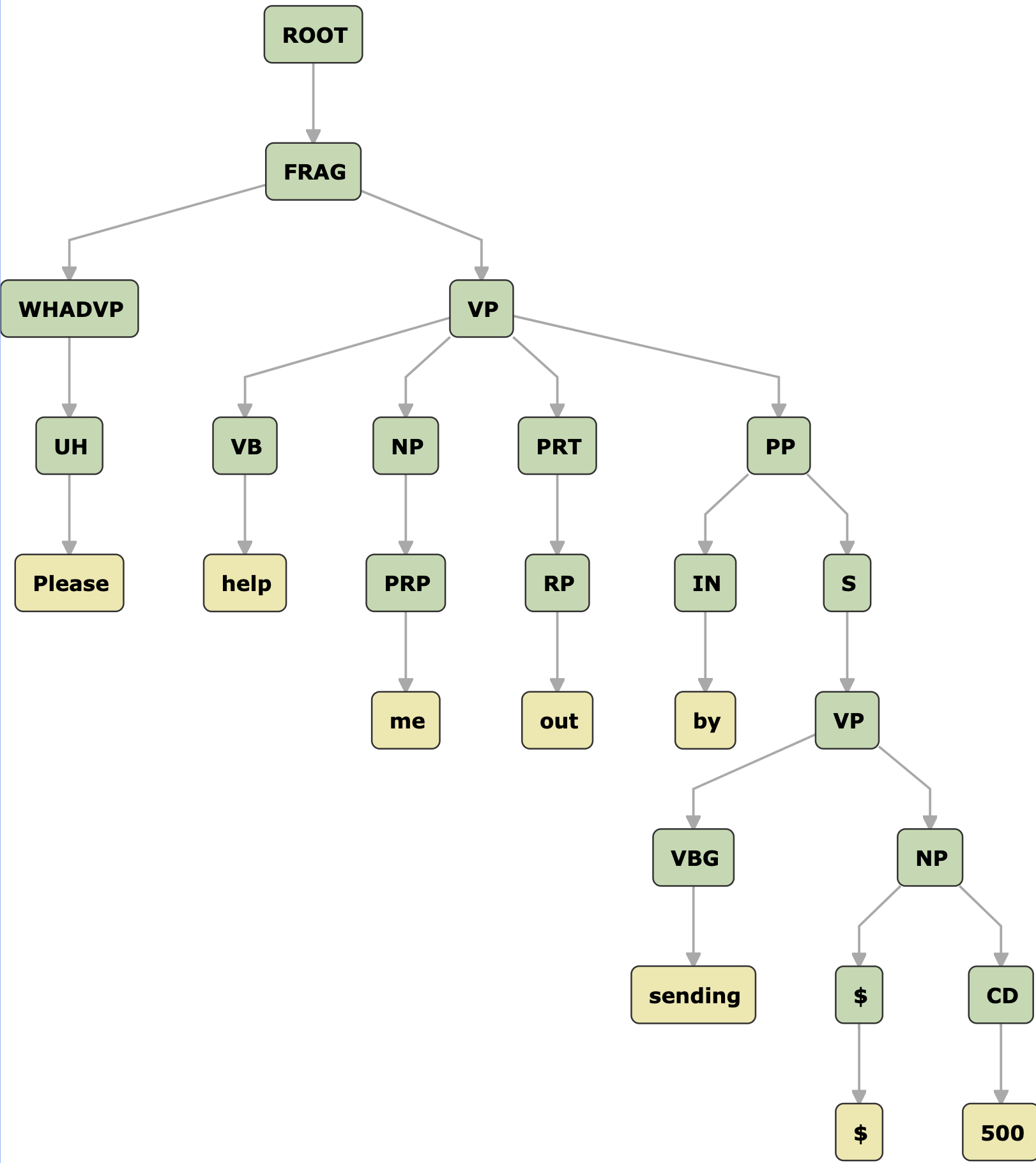}\\ \hline
\textbf{c.~Semantic role labeling:}\\
\mbox{~}\textbf{help:} [V:help][ARG1:me]out[ARGM: by sending \$500]\\
\mbox{~}\textbf{sending:} [V:sending][ARG1:\$500.]\\ \hline
\end{tabular}
\caption{Dependency, Constituency, and Semantic Role Labeling for \textit{Please help me out by sending \$500} }
\label{Fig:Dep-Constit-SRL}
\end{figure}

\noindent
\textbf{a.~Basic Language Tools:}~Detection of the actions associated with asks and framing relies on constituency parses and dependency trees, both taken from Stanford CoreNLP \cite{CoreNLP:2018}.  As shown in Figures~\ref{Fig:Dep-Constit-SRL}a and~\ref{Fig:Dep-Constit-SRL}b, both \textit{help} and \textit{sending} are verbal (VB and VBG, respectively) and thus are extracted as candidate ask actions.  
Arguments are extracted using \textit{semantic role labeling} (SRL) \cite{AllenNLP:2017} (see Figure~\ref{Fig:Dep-Constit-SRL}c), e.g., ARG1 (\$500) is identified as an argument of \textit{sending}.\footnote{Arguments are assigned ask/framing categories mentioned earlier (e.g., finance\_money, personal, credentials).}
The dependency tree may not always yield all possible verbs, so back-off to the constituency tree is sometimes needed. Also, when SRL is not able to populate all arguments, the dependency tree is used to determine arguments for each ask or framing action.

\noindent
\textbf{b.~Thesaurus.com and Lexical Conceptual Structure:}~Two thesaurus-like baselines and an extended verb classification (see Ask Detection webpage) are used to test linguistic and structural constraints on ask/framing detection. The first is a standard but relatively robust thesaurus (thesaurus.com) that yields four lists based on asks (PERFORM, GIVE) and framings (LOSE, GAIN). Examples are shown below (with total verb count in parentheses):

\begin{tabular}{l}
$\bullet$~PERFORM (44): achieve, act, do, execute, perform\\
$\bullet$~GIVE (55): commit, donate, grant, provide\\
$\bullet$~LOSE (41): expend, forefeit, squander, yield\\
$\bullet$~GAIN (53): clean, get, obtain, profit, reap
\end{tabular}

The second baseline is the STYLUS variant of Lexical Conceptual Structure (LCS) \cite{dorr-voss:2018,Dor:18a}, which groups verbs into classes according to syntactic behavior and underlying semantic structure \cite{Levin:1993}. For example, verbs corresponding to GIVE include those in classes such as \textit{Contribute}, \textit{Future Having}, and \textit{Fulfilling}. Some examples are shown below:

\begin{tabular}{l}
$\bullet$~PERFORM (214): ask, bring, execute, rate, redeem\\
$\bullet$~GIVE (81): administer, contribute, donate, furnish\\
$\bullet$~LOSE (615): penalize, stick, punish, ruin\\
$\bullet$~GAIN (49): accept, earn, grab, win
\end{tabular}

Due to their size and coverage, in comparison to thesaurus.com, LCS classes are likely to predict more true positives, but also more true negatives, during ask/framing detection. The benefit and main distinguishing feature of LCS is its extensible organizational structure, which facilitates rapid modifications due to the grouping of similar verbs.  

An extended LCS classification (LCS+) was thus produced from suspected scam/impersonation emails collected by PANACEA team members. Verbs from these emails were tied into particular LCS classes with matching semantic peers and argument structures (one person-day of effort). Modifications and examples are shown below:

\begin{tabular}{l}
$\bullet$~PERFORM (6 del, 44 added): connect, copy, notify\\
$\bullet$~GIVE (no changes): Orig.~LCS covers GIVE verbs\\
$\bullet$~LOSE (174 del, 11 added): deny, forget, surrender\\
$\bullet$~GAIN (no changes): Orig.~LCS covers GAIN verbs
\end{tabular}

\noindent
\textbf{c.~Categorial Variation:}~We incorporate CATVAR (\textit{Categorial Variation Database}) \cite{Nizar:2003} to map between different parts of speech, e.g., \textit{winner}(N) $\rightarrow$ \textit{win}(V). Basic Language Tools (see above) focus primarily on processing verbal parts of speech (VB, VBG, VBD,...) which may miss some asks/framings, 
e.g., \textit{you can reference your gift card} is an implicit ask to examine a gift card associated with the potential victim. CATVAR enables identification of the ask action as \textit{refer}, a PERFORM verb.\footnote{
To our knowledge, no stemming or lemmatization is as comprehensive as CATVAR, which incorporates numerous resources (Brown, NOMLEX, WordNet, etc.), providing multiple variants for any event. For example, \textit{develop} is classified with \textit{developer} (N), \textit{developing} (AJ), \textit{development} (N), \textit{developmental} (AJ), \textit{developmentally} (AV), and several others (a total of 16 variants). The Ask Detection webpage includes a link to CATVAR.}

\noindent
\textbf{d.~Verbal Processing:}~Verbal processing eliminates 
spurious asks containing verb forms such as \textit{sent} or \textit{signing} in \textit{We sent you this email because you're signing up for a new account}. 
Verbal constraints
rule out ask candidates tagged with parts of speech VBG or VBD, thus eliminating \textit{sent} (VBD) and \textit{signing} (VBG) as asks. These constraints do not apply to framing candidates, which may carry any verbal part of speech, e.g., \textit{you won, you are winning}.

\subsection{Application of Structural Knowledge}

\noindent
\textbf{a.~Email structure and links:}~It is 
common for emails to include both HTML and plain text parts with similar semantic and pragmatic content.
Social engineers 
exploit the conventional similarity between HTML and plain text by substituting malicious links and contact information where trusted resources are expected by the user \cite{hadnagy2015phishing}. 
This attack pattern for deception motivates us to focus on the
\texttt{text/html} MIME type when it is available, activating
pre-processing before the linguistic elements of the message are analyzed. 
Processed HTML parts result in text that is line split whenever \texttt{div},  \texttt{p},  \texttt{br}, or  \texttt{ul} tags are encountered for improved sentence splitting. A unique tag is inserted as a placeholder for hyperlinks that may be recovered later as needed. Image tags are replaced with their alt text. Any element that conventionally or canonically indicates styling, scripting, quoting, replying, or a signature is removed.

\noindent
\textbf{b.~Link Positioning:}~Social engineers employ many techniques to entice the potential victim to click on links, including a wide range of different link positionings:
\begin{tabular}{ll}
$\bullet$&Click {\color{blue}\underline{here}}\\
&Ask: PERFORM click({\color{blue}\underline{$<$URL$>$}})\footnotemark\\
$\bullet$&{\color{blue}\underline{Click here}}\\
&Ask: PERFORM click({\color{blue}\underline{$<$URL$>$}})\\
$\bullet$&Get ready to {\color{blue}\underline{vote}} for the best-looking dog.\\
&Ask: PERFORM vote({\color{blue}\underline{$<$URL$>$}})\\
$\bullet$&Contact me. I'm around Mon. ({\color{blue}\underline{jw11@example.com}})\\
&Ask: PERFORM contact({\color{blue}\underline{jw11@example.com}})
\end{tabular}
\footnotetext{{\color{blue}\underline{$<$URL$>$}} refers to the URL embedded in the link.}
In the first three cases above, \textit{Basic} link processing assigns the link to the appropriate ask: the links are embedded in the sentence containing the ask (PERFORM) and its associated action (\textit{click} or \textit{vote}). By contrast, the fourth entails \textit{Advanced} link processing to tie together a link with its corresponding ask-containing sentence, which is separated by intervening material. As we will see shortly, handling both cases increases ask-detection confidence, thus improving the detection of top asks on a per-email basis.

\subsection{Algorithmic Steps and Confidence Score}

Because our approach uses linguistically-motivated rules coupled with structural knowledge, no training data are needed. Algorithmic steps are described below.

\noindent
\textbf{a.~Detect Ask/Framing actions:}~The first step for ask/framing detection is to extract the main action for each clause (recursively) from the dependency tree and the constituency parse shown earlier in Figure~\ref{Fig:Dep-Constit-SRL}a,b. This is achieved first through application of basic language tools and also through application of CATVAR to detect actions that may be implicit in non-verbal forms, such as \textit{reference} (which maps to the PERFORM form \textit{refer}).  Verbal constraints are then applied to
rule out past and progressive actions (VBD/VBG) as asks. If ruled out, the action is considered as a framing candidate. If not ruled out, a priority scheme is applied, attempting to match the action against asks PERFORM and GIVE, in that order, from the lexical resource of interest (the thesaurus or LCS/LCS+). If this fails, an attempt is made to match the action against framings LOSE and GAIN, in that order, using the same lexical resource.

This priority scheme was devised to support overlapping ask actions, e.g., \textit{send} is both a GIVE and  PERFORM in LCS+, but in the context of a clickable link, it is deemed a PERFORM. In this way, \textit{structural knowledge} influences the \textit{linguistic choice} of ask.  Similar overlap exists for framing, e.g., \textit{retrieve} is both a GAIN or a LOSE, depending on the perspective of interest. Given that our application of ask detection is designed for SE interactions, it is assumed that a loss is intended for the potential victim (not for the social engineer); thus, LOSE is tested ahead of GAIN.

\noindent
\textbf{b.~Determine Ask/Framing Arguments:}~Following the detection of an ask or framing action, basic language tools identify the arguments. For example, semantic role labeling identifies \textit{\$500} as ARG1 in the sentence \textit{sending \$500} (see Figure~\ref{Fig:Dep-Constit-SRL}); this becomes an ask argument that is subsequently assigned an ask category as described next.

\noindent
\textbf{c.~Assign Ask/Framing Category:}~Categories are associated with asks and framings, e.g., \textit{sending \$500} yields a GIVE ask with argument \$500, which is in the  finance\_money category. Other examples are shown below:

\begin{tabular}{l}
    $\bullet$~\textit{...using your gift card}: {scam\_gift}\\
    $\bullet$~\textit{Sign-up with your login and password}: {credentials}\\
    $\bullet$~\textit{...confirm with us via this email...}: {personal}\\
\end{tabular}
The categories are hierarchically organized, with a total number of 13 categories.
From this categorization it is possible to deduce the likely goals of a would-be attacker, for use in downstream response generation.\footnote{Although the full description of ask/framing categories is out of scope for this paper, these categories provided hints to the human adjudicator for the generation of our validation set.} 

\noindent
\textbf{d.~Detect Links:}~Links are detected through either basic or advanced link processing and these are associated with ARG1 of the ask (found by basic processing tools). The existence of a link boosts the confidence score for its associated ask. For example, a detached link is found via advanced link processing for: \textit{Contact me. I'm around Mon. ({\color{blue}\underline{jw11@example.com}})}. Here, \textit{me} is associated with the contact email address.

\noindent
\textbf{e.~Apply Confidence Score:}~Application of confidence scores is based on preliminary trial-and-error studies and intuitions gleaned from processing development data. Observations are:  (1) Past tense events are found not to be asks, thus assigned low or 0 confidence;  (2) Non-past-tense events are more prevalently observed to be PERFORM asks if an ask category is specified (e.g., \textit{finance\_money} for \textit{sending \$500} above), thus assigned high confidence (0.8);  (3) The vast majority of asks associated with URLs (e.g., {\color{blue}\underline{jw11@example.com}} tied to \textit{me} above) are found to be PERFORM asks, thus assigned a highest confidence (0.9); (4) GIVE combined with any ask category (e.g., \textit{contribute \$50} above) is less frequently found to be an ask, thus assigned slightly lower confidence (0.75); and  (5) GIVE by itself is even less likely found to be an ask, thus assigned a confidence of 0.6 (e.g., \textit{donate often}). (Automatic confidence scoring, training on actual data, is an area of future work.)

\noindent
\textbf{f.~Select Top Ask:}~Upon completion of the processing above, \textit{Top Ask} selection produces the most important asks at the aggregate level of a single email. This is crucial for downstream processing of the framing and ask (i.e., automatic response generation). Asks are sorted based on their confidence scores, bringing those with the highest scores to the top. Those tied for first place are returned as the ``top asks'' for the email. For example, ``PERFORM contact me ({\color{blue}\underline{jw11@example.com}})'' is returned as the top ask for \textit{Contact me. I'm around Mon. ({\color{blue}\underline{jw11@example.com}})}.

\section{Evaluation Experiments and Results}

This section describes a range of different experiments to demonstrate the impact of  linguistic and structural knowledge on ask/framing detection. It has been argued that the evaluation of dialogue systems is best achieved through comparison to a large set of human-generated responses \cite{Gupta2019}. However, with the intermediate step of ask/framing detection---a well-defined sub-problem of the larger dialogue task in a social-engineering setting---it is possible to achieve a very informative evaluation using a single validation set with straight-forward labels. 

We produce a validation set through human adjudication and correction (by a computational linguist) of initial ask/framing labels automatically assigned by our system to SRL-processed clauses from a held-out test set of 20 emails (472 clauses). The emails contain examples of everyday spam,
targeted attacks by would-be social engineers, and test emails. The resulting validation set is used as a form of \textit{ground truth} (see Ask Detection webpage) against which we measure clause-level precision/recall/F, as described below.

The adjudication task has a throughput of about 50 labeled clauses per hour, for a validation set size of 472 clause-level outputs in one person-day of work. Three types of labels are adjudicated and corrected for each clause-level output to produce the validation set:  (a) ask labels for all asks identified in each email; (b) framing labels for all framings identified in each email; and (c) the top ask (or set of top asks), as determined by the confidence score, for each email---i.e., those considered the most critical, by a human adjudicator, for downstream response generation.


Our experiments focus on the range of true positives/negatives (TP, TN) and false positives/negatives (FP, FN)  that arise with different combinations of linguistic and structural knowledge used for ask/framing detection.\footnote{See TP/TN/FP/FN tallies under Experimental Details on the Ask Detection webpage.} We judge the success of our approach
at the clause level (472 extracted verb-argument excerpts)
using metrics of Precision (P), Recall (R), and F-measure (F).  Test conditions for our experiments are shown in Table~\ref{tab:conditions} for cases 1--6, excluding case 0 (thesaurus baseline). Each column includes all features of the prior column. 
\begin{table}
\centering
\begin{tabular}{|l|l|l|l|l|l|l|}\hline
&1&2&3&4&4&6\\ \hline
Orig LCS Classes    &   Y  &    &   &   &   &  \\ \hline
Class expansions &   Y  &  Y &   &   &   &  \\ \hline
Verbal Processing  &   Y  &  Y & Y &   &   &  \\ \hline
CATVAR         &   Y  &  Y & Y & Y &   &  \\ \hline
Basic Link      &   Y  &  Y & Y & Y & Y &  \\ \hline
Advanced Link        &   Y  &  Y & Y & Y & Y & Y\\ \hline
\end{tabular}
\caption{Experimental conditions for Cases 1--6}
\label{tab:conditions}
\end{table}

Table~\ref{Tab:Results1} provides all experimental results.
\begin{table}[tb]
\begin{small}
\begin{center}
\begin{tabular}{|p{.45in}|r|r|r|r|r|r|r|}\hline
\textbf{Type}&\textbf{TP}&\textbf{TN}&\textbf{FP}&\textbf{FN}&\textbf{P}&\textbf{R}&\textbf{F}\\
\hline
\hline
\multicolumn{8}{|c|}{\textbf{Case 0: Thesaurus Only}}\\
Ask:    & 3 &  392 &  8 &  69 &  0.273 & 0.042 & 0.072 \\
Framing:& 9 &  422 &  25 &  16 &  0.265 & 0.360 & 0.305  \\
TopAsk:& 3 &  411 &  8 &  50 &  0.273 & 0.057 & 0.094  \\
\hline
\hline
\multicolumn{8}{|c|}{\textbf{Case 1: Original LCS Classes}}\\
Ask:&    8 &  378 &  28 &  58 &  0.222 & 0.121 & 0.157  \\
Framing:&  14 &  420 &  30 &  8 &  0.318 & 0.636 & 0.424  \\
TopAsk:& 9 &  409 &  10 &  44 &  0.474 & 0.170 & 0.250  \\
\hline
\hline
\multicolumn{8}{|c|}{\textbf{Case 2: LCS+ Classes}}\\
Ask:*&    \textbf{34} &  365 &  34 &  39 &  0.500 & 0.466 & 0.482  \\
Framing:*& \textbf{15} &  437 &  10 &  10 &  \textbf{0.600} & \textbf{0.600} & \textbf{0.600} \\
TopAsk:& 14 &  401 &  18 &  39 &  0.438 & 0.264 & 0.329  \\
\hline
\hline
\multicolumn{8}{|c|}{\textbf{Case 3: LCS+\&Verbal}}\\
Ask:*&    29 &  384 &  15 &  44 &  0.659 & 0.397 & 0.496  \\
Framing:& \textbf{15} &  437 &  10 &  10 &  \textbf{0.600} & \textbf{0.600} & \textbf{0.600} \\
TopAsk:& 13 &  407 &  12 &  40 &  0.520 & 0.245 & 0.333  \\
\hline
\hline
\multicolumn{8}{|c|}{\textbf{Case 4: LCS+\&Verbal\&CATVAR}}\\
Ask:&    30 &  384 &  15 &  43 &  \textbf{0.667} & \textbf{0.411} & \textbf{0.508} \\
Framing:& \textbf{15} &  437 &  10 &  10 &  \textbf{0.600} & \textbf{0.600} & \textbf{0.600} \\
TopAsk:& 13 &  407 &  12 &  40 &  0.520 & 0.245 & 0.333  \\
\hline
\hline
\multicolumn{8}{|c|}{\textbf{Case 5: LCS+\&Verbal\&CATVAR\&BasicLink}}\\
Ask:&    30 &  384 &  15 &  43 &  \textbf{0.667} & \textbf{0.411} & \textbf{0.508}  \\
Framing:& \textbf{15} &  437 &  10 &  10 &  \textbf{0.600} & \textbf{0.600} & \textbf{0.600} \\
TopAsk:*& 17 &  411 &  8 &  36 &  0.680 & 0.321 & 0.436  \\
\hline
\hline
\multicolumn{8}{|c|}{\textbf{Case 6: LCS+\&Verbal\&CATVAR\&AdvancedLink}}\\
Ask:&    30 &  384 &  15 &  43 &  \textbf{0.667} & \textbf{0.411} & \textbf{0.508}  \\
Framing:& \textbf{15} &  437 &  10 &  10 &  \textbf{0.600} & \textbf{0.600} & \textbf{0.600}  \\
TopAsk:& \textbf{18} &  411 &  8 &  35 &  \textbf{0.692} & \textbf{0.340} & \textbf{0.456}  \\
\hline
\end{tabular}
\end{center}
\caption{\label{Tab:Results1} Impact of combinations of linguistic and structural knowledge on ask/framing detection.
}
\end{small}
\end{table}
McNemar tests were applied to determine statistical significance of performance changes between consecutive cases \cite{McNemar47samplingIndependence}. Asterisked(*) rows have  statistically significant improvements at 5\% level.\footnote{Tested values were correct responses (TP or TN) vs. incorrect responses (FP or FN), significance of change in total error rate.} Statistically significant improvements were found in moving from LCS to LCS+ (for Ask and Framing), in moving from LCS+ to LCS+Verbal  (for Ask), and in moving from LCS+Verbal+CATVAR to Basic Link Processing (for TopAsk).

\section{Results Analysis: Failures and Successes}

Experiments reveal that LCS+ improves detection of both asks and framings, with more than a six-fold increase in F over a standard thesaurus for ask detection (0.482). Higher results are obtained for ask-detection with verbal processing (0.496), and even higher when CATVAR is added (0.508). Two types of structural (link) processing techniques impact the top-ask results, ultimately achieving an F-score of 0.456, four times higher than that of the thesaurus-only baseline (0.094). We highlight certain cases of failure and success:

\textbf{Case 0:} Thesaurus.com yields 3 asks and 9 framings, and identifies 3 top asks. 
F-scores are modest, 0.30 or less for asks, framing, and top asks, e.g., \textit{take part in this conference} is correctly identified as PERFORM, but \textit{sign up} is missed.

\textbf{Case 1:} Original LCS Classes yield more correct asks and framings in comparison to Case 0, with more than a two-fold improvement in F for asks (from 0.072 0.157) and a 40\% increase for framing (from 0.305 to 0.424). Unlike Case 0, LCS classes detect \textit{sign up} as a PERFORM ask. However, false positives increase, particularly for asks (from 8 to 28), e.g., LCS incorrectly deems \textit{we value your participation} to be a PERFORM ask. Correct top asks climb from 3 to 9, a three-fold improvement, e.g., LCS assigns \textit{Did you send money?} as a top ask, where Case 0 fails to do so. 

\textbf{Case 2:} LCS+ Classes increase correct asks from 8 to 34, and increase false positives from 28 to 34, over the original LCS with overall F-score improvements. For example, LCS+ correctly identifies \textit{contact me} as a PERFORM, where the original LCS does not, but also incorrectly identifies \textit{you would rather not receive} as a PERFORM. False positives for framing are reduced over the original LCS (30 to 10), e.g., the original LCS, but not LCS+, incorrectly detects LOSE for \textit{dog could be a star}. 
False positives for correct top asks increase from 10 to 18, e.g., \textit{we sent this email} is deemed a PERFORM, mitigated by verbal processing (below).
    
\textbf{Case 3:} Verbal Processing is coupled with LCS+, to filter past and progressive forms of predicted asks, thus eliminating sentences like \textit{we sent this email} as an ask. This verbal constraint significantly reduces false positives for asks (34 to 15) over LCS+ alone.\footnote{Framing is not subject to verbal constraints; thus the number of framings does not change (cases 2--6 remain the same for framing).}  Correct top asks drop slightly (14 to 13) but false positives reduce significantly (18 to 12), e.g., \textit{is being sent} is eliminated as a top ask. Although verbal constraints do not increase correct ask types, fewer false positives enables more effective downstream response generation. Verbal constraints combined with structural processing (below) yield more correct top asks than LCS+ alone, with F increasing about 21\% (14 to 17) for basic link processing, and 5\% (17 to 18) for advanced link processing. 

\textbf{Case 4:} CATVAR raises true positives for asks slightly (29 to 30) without increasing false positives (15). Top asks remain at 13 and framings do not change. CATVAR adds one true ask---a case where \textit{reference}(N) is mapped to \textit{refer}(V). One explanation for CATVAR's unanticipated low impact is that social engineers generally do not employ verb-nominal combinations. Example emails developed among PANACEA team members include  multi-word constructions such as \textit{You \textbf{emerge winner} of \$1M}. This is a place where CATVAR would have mapped to the verb \textit{win}. However, true SE data are more likely to use phrases such as \textit{You have won}.

\textbf{Cases 5 and 6:} Basic/Advanced Link Processing leverages structural knowledge, which does not impact correct asks (as expected), but directly impacts true and false positives for top asks. Basic Link processing increases correct top asks significantly (13 to 17) and also significantly reduces false positives (12 to 8), yielding significant F-score improvement (0.333 to 0.436). For example, basic processing eliminates \textit{fail to bring} as an ask and advanced processing adds one more correct top ask (18): \textit{open a tutorial}.

\section{Related Work}

Security in online communication is a challenging problem, due to several issues:
(1) the attacker's speed outpaces the ability of defenders to maintain indicators \cite{Zhang2006PhindingPE}; (2) the quality of phishing sites are high enough that users ignore alerts \cite{Egelman:2008:YWE:1357054.1357219}; (3) User training falls short as users quickly forget the material and fall prey to previously studied attacks \cite{caputo2013going}; and (4) defensive system maintainers may not always take into account the context, motivations, and socio-economic status of the targeted user \cite{Oliveira2017-phishing}. 

Numerous studies \cite{bak2008,kar2006} 
have demonstrated human susceptibility to SE attacks. 
Moving from bots that detect SE attacks to those that produce ``natural sounding'' responses and engage the attacker to elicit identifying information 
is the next advance in this arena. 
We take the first step in our work on ask/framing detection, to be used for downstream processing by conversational agents (CAs).
 
CAs suffer from the problem of generating dull and generic responses \cite{gao2019neural,santhanam2019survey}. To address this issue, 
topic models are used to produce focused responses augmenting existing neural based approaches to CAs \cite{dziri2018augmenting}. 
Targeted responses employ self-disclosure as a strategic approach for building an engaging conversation \cite{RavichanderB18}. 
For SE detection, 
topic models  \cite{Bhakta:2015} and NLP of conversations \cite{Yuki:2016} are leveraged. However, all of these approaches are limited to a pre-defined set of topics, constrained by the training corpus. 

Prior work on detecting and predicting persuasion in discussions \cite{mckeown:aaai-2018} leverages argument structure as a determining factor for judging when a persuasive attempt might be successful. This work has been adopted for (subreddit) forum discussions specifically dedicated to changing opinions (ChangeMyView subreddit). 
Our work is related to this but aims to achieve effective dialogue for countering (rather than adopting) persuasive attempts.


Many approaches address issues above via machine learning. Structured knowledge representations of scripts embedded in latent space are used to detect and compare similar events  \cite{li-goldwasser-2019-encoding}. This is useful for determining whether an email resembles a password reset email typically sent from an organization's IT department. Comparison of multiple responses in a chatbot setting has been shown to improve correlation of evaluation metrics with human judgement \cite{prakhar-2019-open-domain-dialogue}. This can be used to select a chatbot model that performs better in an open domain. However, unlike our approach, these approaches require extensive model training.

Closest to our work is text-based semantic analysis for detection of SE attacks \cite{kimcatch}. Our work differs in that it focuses not just on \textit{detecting} an attack, but on \textit{engaging} with an attacker in ways that leverage \textit{asks} and \textit{framings}. Whereas a chatbot might be employed to warn a potential victim that an attack is underway, e.g., based on malicious content, our chatbots are designed to communicate with a social engineer in ways that elicit identifying information. Extraction of asks/frames supports generation of responses to the attacker---with an end goal of eliciting attributable information about the attack in order to identify them.


\section{Conclusions and Future work}

We appeal to foundational notions of \textit{ask} and \textit{framing} to characterize SE attacks 
for downstream 
response generation. We conduct experiments to determine the impact of linguistic and structural knowledge on ask and framing detection, coupled with top ask selection.  We make use of 
parts of speech, 
categorial variations, and verb classes informed by lexical-conceptual structure to yield significant precision and recall improvements for ask and framing detection. We also apply structural clues such as link detection for 
improved recall of ``top asks'' for individual emails. 

As noted earlier, categorial variations (CATVAR) provide only small improvements to ask-detection performance.
Certain types of multi-word expressions merit further investigation within the SE domain, most notably light verb constructions, which are not yet accommodated in our approach: \textit{take note} vs. \textit{notice}, \textit{take into account} vs. \textit{account for}, etc. 

We also expect to enrich our LCS+ classes through the addition verbs that do not currently appear in LCS+. For example, several thesaurus verbs for GIVE are missing from the LCS+ verbs for GIVE: \textit{allow, commit, endow, sell}. A future goal is to use a class-based combination of multiple resources to improve ask-detection performance compared to using either resource alone.

We are conducting experiments to demonstrate the utility of ask detection in an extrinsic evaluation of a conversational agent. We hypothesize that ask detection enables more targeted responses than would otherwise be generated and that the degree of sophistication of the ask-detection component correlates with the reliability of response generation. Example conversations are included on the Ask Detection page.

\paragraph{Acknowledgments:} 
This work was supported by DARPA through AFRL Contract FA8650-18-C-7881 and through Army Contract W31P4Q-17-C-0066. All statements of fact, opinion or conclusions contained herein are those of the authors and should not be construed as representing the official views or policies of DARPA, AFRL, Army, or the U.S. Government.

\fontsize{9.5pt}{10.5pt}
\selectfont
\bibliography{AAAI-DorrB.9586}
\bibliographystyle{aaai}

\end{document}